\pdfoutput=1

\documentclass[11pt]{article}

\usepackage[final]{acl}

\usepackage{times}
\usepackage{latexsym}
\usepackage{tablefootnote}

\usepackage{multirow}
\usepackage{graphicx}
\usepackage[T1]{fontenc}

\usepackage[utf8]{inputenc}

\usepackage{microtype}

\usepackage{inconsolata}

\title{TM-TREK at SemEval-2024 Task 8: Towards LLM-Based Automatic Boundary Detection for Human-Machine Mixed Text}

\author{Xiaoyan Qu \and Xiangfeng Meng \\
        Samsung R\&D Institute China-Beijing \\
         \{xiaoyan11.qu, xf.meng\}@samsung.com}

\begin{document}
\maketitle
\begin{abstract}
With the increasing prevalence of text generated by large language models (LLMs), there is a growing concern about distinguishing between LLM-generated and human-written texts in order to prevent the misuse of LLMs, such as the dissemination of misleading information and academic dishonesty. Previous research has primarily focused on classifying text as either entirely human-written or LLM-generated, neglecting the detection of mixed texts that contain both types of content. This paper explores LLMs' ability to identify boundaries in human-written and machine-generated mixed texts. We approach this task by transforming it into a token classification problem and regard the label turning point as the boundary. Notably, our ensemble model of LLMs achieved first place in the 'Human-Machine Mixed Text Detection' sub-task of the SemEval'24 Competition Task 8. Additionally, we investigate factors that influence the capability of LLMs in detecting boundaries within mixed texts, including the incorporation of extra layers on top of LLMs, combination of segmentation loss, and the impact of pretraining. Our findings aim to provide valuable insights for future research in this area.
\end{abstract}

\section{Introduction}
Large language models (LLMs), particularly since the debut of ChatGPT, have made significant advancement and demonstrated the ability to produce coherent and natural-sounding text across a wide range of applications. However, the proliferation of generated text has raised concerns regarding the potential for misuse of these LLMs. One major issue is the tendency of LLMs to produce hallucinated content, resulting in text that is factually inaccurate, misleading, or nonsensical. Inappropriate utilization of LLMs for text generation purposes, such as in news articles \cite{2019Defending}, social media posts \cite{2020TweepFake}, and app reviews \cite{2019Release}, can propagate misinformation and influence public perceptions. Furthermore, the use of machine-generated text can also facilitate academic dishonesty. Therefore, accurately distinguishing between human-authored and machine-generated texts is crucial in order to address these challenges effectively.

The majority of existing studies addressing this challenge have approached it as a machine-generated text classification problem, aiming to determine whether a given text is generated by LLMs or not. However, this approach assumes that the text is either completely machine-generated or entirely human-written. With the increasing collaboration between humans and AI systems, mixed texts containing both human-authored and machine-generated portions have emerged as a new scenario that simple machine-generated text classification methods cannot effectively address \cite{dugan2023real}. Therefore, a more nuanced approach to machine-generated text classification for mixed texts is necessary.

This study addresses the challenge of token-level boundary detection in mixed texts, where the text sequence starts with a human-written segment followed by a machine-generated portion. The objective is to accurately determine the transition point between the human-written and LLM-generated sections. To achieve this, we frame the task as a token classification problem, thus the turning point of the label sequence will be the boundary. Through experiments utilizing LLMs that excel in capturing long-range dependencies, we demonstrate the effectiveness of our approach. Notably, by leveraging an ensemble of multiple LLMs to harness the robust of the model, we achieved first place in Task 8 of SemEval'24 competition.

Furthermore, we explore factors that impact the effectiveness of LLMs in boundary detection, including the integration of additional layers on top of LLMs, the combination of segmentation loss and pretraining techniques. Our experiments indicate that optimizing these factors can lead to significant enhancements in boundary detection performance. 

The main contribution of this papers include:

1) We explore LLMs' capability to detect boundaries within human-machine mixed texts, compare the performance of various LLMs, and present a benchmark based on the new released data set. And we rank 1st in the corresponding SemEval'24 competition \cite{semeval2024task8}.

2) We examine factors that impact boundary detection in mixed texts, including additional layers on top of LLMs, introduce of segment loss functions, and pretraining technique. We aim to provide valuable insights for future research in this field.

\section{Related Work}
Previous research has predominantly focused on machine-generated text classification \cite{2022Machine,2020Automatic}, where the text is attributed to either human authors or large language models. The objective is to determine whether a given text is human-written or specifically generated by a particular LLM. These studies can be classified into two main categories: metric-based methods and model-based methods. Metric-based approaches leverage metrics such as word rank, predicted distribution entropy, and log-likelihood \cite{2023Detectgpt, 2019GLTR, 2023gptwho}. On the other hand, model-based methods involve training models on labeled data \cite{2022coco}. However, these methods are not directly applicable to boundary detection for mixed human-machine texts.

Recently, there are a few works investigating the detection of mixed human-machine text. These texts consist of both human-written and machine-generated content, and the objective is to accurately identify the boundary between these two segments. \citet{dugan2023real} delved into the human ability to discern boundaries between human-written and machine-generated text. Their study revealed significant variations in annotator proficiency and analyzed the impact of various factors on human detection performance. \citet{zeng2023towards} were the first to formalize the task as identifying transition points between human-written and AI-generated content within hybrid texts, and they examined automated approaches for boundary detection. 

One limitation of these studies is that the transitions typically occur between sentences rather than at the word level. This paper aims to address the token-level boundary detection of mixed texts.

\section{Methodology}
\subsection{Task Formulation}
The task is presented as a sub-task 'Subtask C: Human-Machine Mixed Text Detection' in SemEval'24 Task 8\footnote{\url{https://github.com/mbzuai-nlp/SemEval2024-task8/tree/main}}. The task is defined as follows: for a hybrid text $<w_1, w_2, ..., w_n>$ with a length of $n$ that includes both human--written and machine-generated segments, the objective is to determine the index $k$, at which the initial top $k$ words are authored by humans, while the subsequent are generated by LLMs. The evaluation metric for this task is Mean Absolute Error (MAE). It measures the absolute distance between the predicted word and the actual word where the switch between human and machine occurs.

We transform the task of boundary detection in mixed texts into a token classification task, aligning it with the competition baselines. Token classification involves assigning a label to each token within a text sequence. In boundary detection tasks, we utilize two labels to indicate whether each token was written by humans or generated by LLMs. By predicting the label of each token in the text sequence, we can identify the specific word that signifies the boundary between the human-written and machine-generated portions of the text.

\subsection{LLM based Boundary Detection}
\begin{figure*}[htbp]
\centering
\includegraphics[scale=0.55]{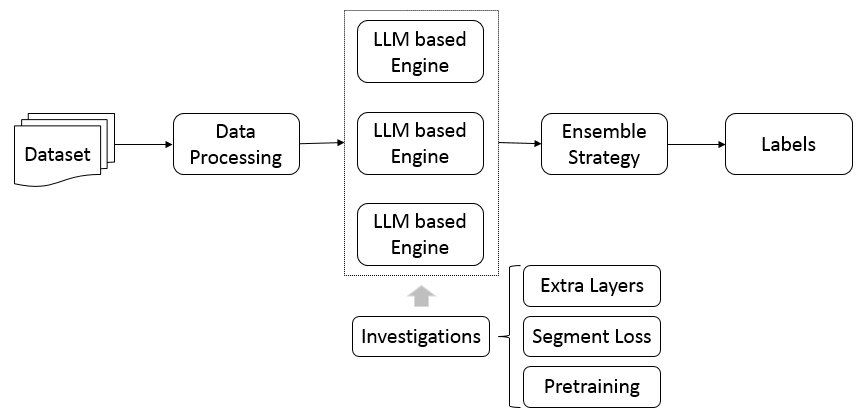}
\caption{Framework of this paper}
\label{figure:framework}
\end{figure*}

We explores LLMs' capability to detect boundaries for human-machine mixed texts. The framework of this paper is shown in Figure \ref{figure:framework}. Given the labeled dataset containing boundary indices, we first map these indices to assign each token a label denoting whether it originates from human writing or LLM generation. Subsequently, we harness the capabilities of LLMs by fine-tuning them for the task of classifying each token's label. To enhance performance further, we employ an ensemble strategy that consolidates predictions from multiple fine-tuned models. Additionally, we explore various factors that impact the effectiveness of LLMs in boundary detection.

\subsubsection{LLMs Supporting Long-range Dependencies}
Our objective is to facilitate boundary detection in long text sequences, thereby necessitating the utilization of LLMs capable of handling long-range dependencies. Within this study, we investigate the performance of Longformer, XLNet, and BigBird models on boundary detection.

\paragraph{Longformer} \cite{2020Longformer} utilizes a combination of global attention and local window-based attention mechanisms, which enables the model's capability to capture both short-range and long-range dependencies effectively.

\paragraph{XLNet} \cite{2019XLNet} introduces a new training objective called permutation language modeling, and considers all possible permutations of the input tokens during training. This allows XLNet to capture bidirectional context more effectively and mitigate the limitations of autoregressive models.

\paragraph{BigBird} \cite{2020Big} introduces a novel sparse attention mechanism that allows the model to scale to longer sequences while maintaining computational efficiency.

\subsubsection{Exploration of Potential Factors}
Except for the direct usage of LLMs, we investigate factors that may influence the capability of LLMs in boundary detection task.

\paragraph{Extra Layers on top of LLMs} While LLMs demonstrate remarkable proficiency in comprehending semantics and generating coherent text, the addition of supplementary layers on top of LLMs has the potential to yield further improvements for downstream tasks. Therefore, we evaluate the impact of additional layers, such as LSTM and CRF, when integrated with LLMs, to ascertain their potential contributions to enhancing performance in boundary detection tasks.

\paragraph{Segment Loss Function} Token classification involves assigning specific categories to individual tokens within a text sequence based on their semantic content. Typically, evaluation metrics gauge the average accuracy of category assignments across all tokens. However, in the context of boundary detection, the labels of tokens situated at or in proximity to the boundary hold greater significance. To bridge this gap, we introduce loss functions capable of assessing segment accuracy for both the human-written and machine-generated segments, such as the dice loss function. These loss functions, commonly utilized in image segmentation tasks that entail dividing an image into distinct segments, are anticipated to enhance performance in boundary detection tasks.

\paragraph{Pretrain and Fine tune} Within the SemEval'24 competition, a total of 4,154 cases are presented for this task. The remaining sub-tasks revolve around human-machine text classification and aim to classify a given text into either human-generated or LLM-generated. A natural idea is to initiate pretraining utilizing the text classification data, followed by fine-tuning on the bounary detection data to enhance the model's overall generalization capability.

Two distinct pretraining approaches are employed. In the Pretrain 1, human-written texts and machine-generated texts are concatenated to form a new boundary detection dataset in sentence level. Within this novel dataset, boundaries are identified at the juncture where human-written and machine-generated sentences merge. Models are trained initially on the sentence-level dataset and subsequently fine-tuned using the 4,154 cases provided. In the Pretrain 2, a binary text classification model incorporating both an LLM and a linear layer atop the LLM is initially trained. Subsequently, the weights of the LLM are utilized for fine-tuning in the boundary detection task. 

\section{Experiments}
\subsection{Dataset}
The data is an extension of the M4 dataset \cite{wang2023m4}. It consists of 3,649 train cases and 505 dev cases, each with text content and gold boundary index. The boundary index denotes the position of the word split caused by a change, with white space serving as the delimiter. Data statistics are shown in Table \ref{table:data_statistics}.

\begin{table}
\centering
\begin{tabular}{lll}
\hline
 & \textbf{Training data} & \textbf{Dev data} \\
\hline
Number & 3649 & 505 \\
Average Length & 263 & 230 \\
Max Length & 1397 & 773 \\
Average Index & 71 & 68 \\
\hline
\end{tabular}
\caption{\label{table:data_statistics}Statistics of the boundary detection data set}
\end{table}

In addition, SemEval'24 Competition also presented two datasets related to binary and multi-way text classification tasks. We investigate whether these additional datasets could potentially enhance boundary detection performance by pretraining techniques. The details of the two supplementary datasets are shown in Table \ref{table:another_2_data_statistics}.

\begin{table}
\centering
\begin{tabular}{lll}
\hline
 & \textbf{Train} & \textbf{Dev} \\
\hline
binary text classification & 119,757 & 5,000 \\
multi-way text classification & 71,027 & 3,000 \\
\hline
\end{tabular}
\caption{\label{table:another_2_data_statistics}Statistics of two other data sets for human-machine text classification tasks}
\end{table}

\subsection{Experimental Results}
\subsubsection{Performance of different LLMs}
We investigate three LLMs renowned for their ability to handle long-range dependencies: Longformer, XLNet, and BigBird. We exclusively employ the large versions of these models. Moreover, as a benchmark for the competition, we utilize Longformer-base. The performance metrics of these four models are outlined in Table \ref{table:single_model_comparison}.

As depicted in Table \ref{table:single_model_comparison}, we observe that Longformer-large outperforms Longformer-base, owing to its increased parameter count. Among the four algorithms, XLNet achieves the best performance, with an MAE of 2.44. This represents a reduction of 31.84\% compared to Longformer-large and a substantial 58.71\% decrease compared to BigBird-large. One potential explanation is that the consideration of all possible permutations of input tokens during training help XLNet capture bidirectinal context more effectively.

\begin{table}
\centering
\begin{tabular}{lll}
\hline
\textbf{Method} & \textbf{MAE} \\
\hline
Longformer (baseline) & 4.11 \\
Longformer-large & 3.58 \\
XLNet-large & \textbf{2.44} \\
BigBird-large & 5.91 \\
\hline
\end{tabular}
\caption{\label{table:single_model_comparison}Performance of varied LLMs}
\end{table}

The winning approach in SemEval'24 Competition is founded on ensembles of 2 XLNet with varied seeds. It involves a simple voting process of the output logits from the diverse XLNet models. As shown in Table \ref{table:XLNet_with_ensemble}, the voting strategy results in a decrease in MAE from 2.44 to 2.22.

\begin{table}
\centering
\begin{tabular}{lll}
\hline
\textbf{Method} & \textbf{MAE} \\
\hline
XLNet-large & 2.44 \\
XLNet-large vote \tablefootnote{The approach that ranks 1st in sub-task C leaderboard} & 2.22 \\
\hline
\end{tabular}
\caption{\label{table:XLNet_with_ensemble} Performance of multiple XLNet ensembles}
\end{table}

\subsubsection{Performance of LLMs with extra layers}
We select Longformer-large as our baseline model and examined the impact of incorporating extra LSTM, BiLSTM, and CRF layers\citep{huang2015bidirectional} on boundary detection. The experimental results are detailed in Table \ref{table:Longformer_with_lstm}. Integration of LSTM and BiLSTM layers with Longformer leads to significant improvements, with a decrease in MAE by 10.61\% and 23.74\%, respectively. Conversely, the addition of a CRF layer to Longformer-large yields unsatisfactory results. One plausible explanation could be the lack of clear dependencies between the two labels (0 and 1) in the token classification task, unlike in tasks such as named entity extraction.

\subsubsection{Performance with segment loss functions}
We investigate the impact of employing segment loss functions commonly utilized in image segmentation \cite{2020A} on boundary detection. The selected loss functions consist of BCE dice loss, Jaccard loss, Focal loss, Combo loss and Tversky loss. Additionally, we introduce a novel loss function BCE-MAE by simply adding BCE and MAE.

\begin{table}
\centering
\begin{tabular}{lll}
\hline
\textbf{Method} & \textbf{MAE} \\
\hline
Longformer-large & 3.58 \\
Longformer-large + LSTM & 3.20 \\
Longformer-large + BiLSTM & 2.73 \\
Longformer-large + CRF & 5.86 \\
\hline
\end{tabular}
\begin{small}
\caption{\label{table:Longformer_with_lstm}Performance of adding extra layers}
\end{small}
\end{table}

We utilize Longformer-large as the baseline, which by default adopts the binary cross-entropy loss. We explore the impact of adjusting the loss functions and present the results in Table \ref{table:Longformer_with_cv_loss}. Among these variations, BCE-dice loss, Combo loss, and the BCE-MAE loss demonstrate superior performance compared to the default BCE loss.

Both BCE-dice loss and Combo loss are initially designed to integrate binary cross-entropy and Dice loss using different weighting schemes to enhance the performance of binary image segmentation. Dice loss serves as a metric for assessing the overlap between the predicted segmentation and the ground truth mask. The introduction of the Dice loss enables a balance between segmentation accuracy and token-wise classification accuracy, resulting in anticipated performance improvements. Compared to the benchmark, the MAE decreases by 12.29\% and 13.69\%, respectively. 

The BCE-MAE loss incorporates MAE loss during the training stage, aligning with the evaluation metric used in the competition. As anticipated, the MAE metric decreases by 16.48\%.

\begin{table}
\centering
\begin{small}
\begin{tabular}{cc||cc}
\hline
\textbf{Method} & \textbf{MAE} & \textbf{Method} & \textbf{MAE} \\
\hline
base (BCE loss) & 3.58 & Dice loss & 3.80 \\
BCE-dice loss & \textbf{3.14} & Jaccard loss & 3.60 \\
Focal loss & 3.40 & Combo loss & \textbf{3.09} \\
Tversky loss & 3.69 & BCE-MAE loss & \textbf{2.99} \\
\hline
\end{tabular}
\caption{\label{table:Longformer_with_cv_loss}Performance of different segment loss functions}
\end{small}
\end{table}

\subsubsection{Performance of LLMs with pretraining}
For both two pretraining approaches, we employ three different settings: directly utilizing Longformer-large for pretraining and fine-tuning; and incorporating additional LSTM and BiLSTM layers, respectively. Table \ref{table:Longformer_with_pretraining} presents the results.

In Pretrain 1, the datasets from the other two subtasks are concatenated to create a new dataset in which the segmentation is on a sentence level, similar to previous studies. Sentence-level boundary detection is akin to token-level boundary detection. So the pretraining of sentence-level data is anticipated to obtain extra gains. The table indicates that simply employing Longformer-large with pretraining reduces the MAE from 3.58 to 3.26. By incorporating LSTM and BiLSTM layers, the MAE is further reduced to 2.85 and 2.84, respectively. 

In Pretrain 2, we initially utilize Longformer to classify whether a given text is human-written or machine-generated. When the pretrained Longformer is fine-tuned directly on boundary detection, only inserting a new linear layer yields poor performance. However, with the inclusion of additional LSTM and BiLSTM layers, it can achieve comparable performance to that of Pretrain 1, reaching 3.04 and 2.72, respectively. 

The results of the two pretraining approaches indicate that pretraining on either sentence-level boundary detection or the binary human-machine text classification task can enhance LLMs' capability to detect token-wise boundaries in mixed texts.

\begin{table}
\centering
\begin{tabular}{lll}
\hline
\textbf{Pretraining} & \textbf{Method} & \textbf{MAE} \\
\hline
No pretrain & Longformer-large & 3.58 \\ 
\hline
\multirow{3}{*}{Pretrain 1} & Longformer-large & 3.26 \\
 &  + LSTM & 2.85 \\
 &  + BiLSTM & 2.84 \\
\hline
\multirow{3}{*}{Pretrain 2} & Longformer-large & 68.52  \\
 &  + LSTM &  3.04 \\
 &  + BiLSTM & 2.72 \\
\hline
\end{tabular}
\begin{small}
\caption{\label{table:Longformer_with_pretraining} Performance of Longformer-large with pretraining}
\end{small}
\end{table}

\section{Conclusion}
\label{sec:conclusion}
This paper introduces LLM-based methodology for detecting token-wise boundaries in human-machine mixed texts. Through an investigation into the utilization of LLMs for boundary detection, we have achieved optimal performance by leveraging an ensemble of XLNet models in the SemEval'24 competition. Furthermore, we explore factors that could affect the boundary detection capabilities of LLMs. Our findings indicate that (1) loss functions considering segmentation intersection can effectively handle tokens surrounding boundaries; (2) supplemental layers like LSTM and BiLSTM contribute to additional performance enhancements; and (3) pretraining on analogous tasks aids in reducing the MAE. This paper establishes a state-of-art benchmark for future researches based on the new released dataset. Subsequent studies aims to further advance the capabilities of LLMs in detecting boundaries within mixed texts.

\bibliography{custom}
\appendix
\end{document}